# An Optimised Brushless DC Motor Control Scheme for Robotics Applications

Nilabha Das[1], Laxman Rao S. Paragond[2], Balkrushna H. Waghmare[3]

[1]*Department of Electrical and Electronics Engineering, Manipal Institute of Technology, India*

[2] *Department of Electrical and Electronics Engineering, Manipal Institute of Technology, India*

[3]*ARTPARK, Indian Institute of Science, India*

[1]`nilabha.das@learner.manipal.edu;` [2]`laxman.sp@manipal.edu;` [3]`balkrushna@artpark.in`

*Abstract*— *This work aims to develop an integrated control strategy for Brushless Direct Current Motors for a wide range of applications in robotics systems. The controller is suited for both high torque - low speed and high-speed control of the motors. Hardware validation is done by developing a custom BLDC drive system, and the circuit elements are optimised for power efficiency.*

## I. INTRODUCTION

Brushless DC Motors, popularly known as BLDCs, are popular in many robotics' applications for their high density of power and high-speed capabilities. These motors have been found to be used in advanced robots, which require precision position and torque control, and in drones and crawlers, which require high speed and smooth transient response. This work aims to design a controller for a wide range of applications and a circuit selection method enabling high-efficiency drives for BLDCs and testing them in real-life situations.

## II. LITERATURE REVIEW

In the excitation of a three-phase BLDC motor, which commonly is achieved using a six-switch three-phase (SSTP) voltage source inverter, only two of the three-phase windings are conducting at a time, and the non-conducting phase carries the back EMF. Exploring this feature, many indirect position detection methods, which sense the back EMF from the floating phase, have been reported in the literature without shaft-mounted position sensors [1][2]. Brushless drives for unmanned aerial vehicles were investigated in [3], and a comparative study of BLDC drives for industrial applications was carried out in [4]. These reported approaches require all three motor terminal voltages to be sampled at all instants of operation. One well-known method is to use filters to extract rotor position information. This is then utilised by the microcomputer to generate the required switching pulses in sequence to control the motor's movement. For low-speed operation, the back-EMF integrating method [5] is a technique applying the principle that integration is constant from Zero Crossing Point (ZCP) to 30˚. The advantage is that the operation of the main processor decreases, as it is not necessary to calculate an additional conversion point during the switching action. A method to utilise the complete DC Bus voltage was proposed in [6], allowing 100% PWM duty cycle operation. Methods for reducing power loss in the inverter stage and increasing the efficiency of the controllers were investigated in [7].

## III. METHODOLOGY

An easy-to-implement and computationally effective method of phase commutation is the BEMF sensing method; in a three-phase BLDCM, only two out of the three-phase windings are energised at an instant of time, and the third, floating phase carries the back-EMF. This back-EMF signal can be synchronously sampled from this phase into the microcontroller. One of the drawbacks of this method lies in the fact that the amplitude of the BEMF signals is a function of motor speed and is almost negligible at lower speeds. Thus, a common practice is to start the motor in an open loop, and after accelerating the rotor towards a sufficient speed where the BEMF signals are sufficient to be sensed, the closed-loop control commences. This method can thus only be used in situations where the operating region of the motor is limited to very high speeds and is not suited for low-speed and high-torque applications, which comprise a major fraction of the use cases of BLDCMs. Fig. 1 shows the zero crossing and commutation points with reference to the BEMF signals:

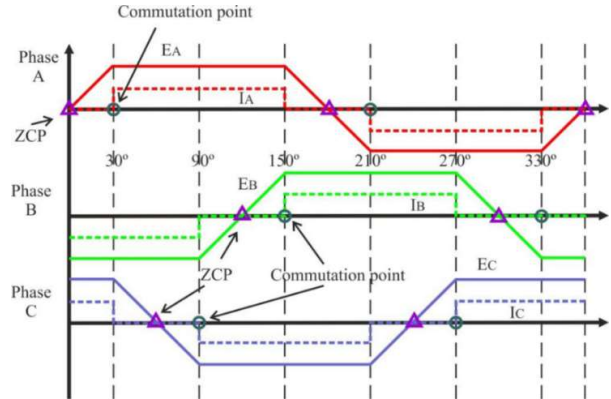

Fig. 1 Commutation points and BEMF of a BLDC Motor

*A. Controller Design*

A method is proposed to ensure optimal performance in both low-speed and high-speed regions using easy-to-implement BEMF sensing methods, which can be achieved by adding a few external components.

In the startup and low-speed region for speed sensing of the BLDC motor, the indirect sensing technique of BEMF voltage integration [5] is used. The integration is based on the concept of area under the curve, where the BEMF voltage area of the floating phase will be approximately the same for low and high speeds. In theoretical terms, this method could work at near zero speeds and is independent of BEMF amplitude and, thus, motor speed. Fig. 2 shows the integrator approach:

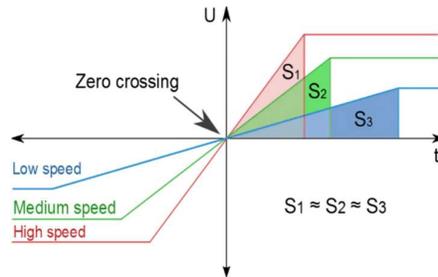

Fig. 2 BEMF integrator across a range of speeds [5]

To function nearly like an ideal integrator, a low-pass filter is employed, which inevitably results in an output phase delay that varies with motor speed. The phase delay that the filter introduces is at maximum $\Pi/2$ and changes with the back EMF frequency, which is determined by the motor speed. If this phase-delay characteristic is not accounted for, the phase-commutation timing will be wrong. It is necessary to fix the phase delays in order to operate a motor satisfactorily at low speeds. The phase delay at a specific frequency can be computed once the filter is designed. An iterative method is employed to create a lookup table for the corrected timing values. Cost savings can be further increased by coupling the sensing circuit with a single-chip microprocessor or DSP for speed control. This improves the operating range and motor efficiency.

At sufficiently high speeds, the BEMF, $E_a, E_b$ and $E_c$ of the three phases have an amplitude that is high enough to be sampled for phase commutation; thus, the integrator stage is not required, and the zero-cross detection on the floating phase can be used to control the motor. In most applications, this method samples the floating phase at the PWM "off" state, which mandates a duty cycle of less than 100%. It is desirable, however, to run the motor at 100% duty cycle to completely utilise the DC-link voltage $V_{bus}$. To overcome this constraint, a method to sample the BEMF at the PWM "on" state is utilised [6]. Fig. 3 shows the circuit analysis of the "on" time sampling method:

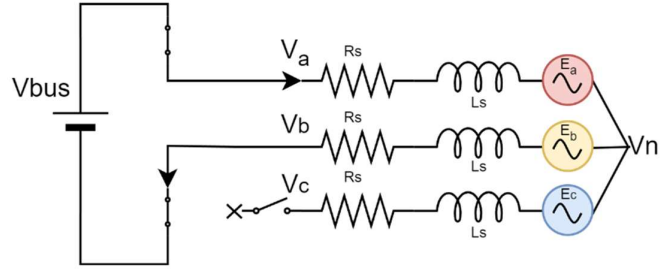

Fig. 3 Sampling the BEMF in PWM "on" time.

By analysing the circuit, $V_n$, the virtual neutral point voltage, according to eq. 1 is:
$$V_n = \frac{V_{bus}}{2} - \frac{E_a + E_b}{2} \tag{1}$$
Since the motor represents a balanced 3-phase machine, according to eq. 2:
$$E_a + E_b + E_c = 0 \tag{2}$$
Thus, the neutral voltage is found out from eq. 3 as:
$$V_n = \frac{V_{bus}}{2} - \frac{3}{2} E_c \tag{3}$$
Therefore, by comparing the floating phase voltage with a reference $V_{bus}/2$ during the PWM "on" interval, the zero-crossing detection can be carried out at a 100% duty cycle. At startup and low speed, the BEMF integrator approach for single-phase sampling through the integrating low pass filter, is selected by P_A MUX, GND reference is provided by REF MUX, and the lookup table provides the corrected timing sequence. Upon reaching sufficient BEMF amplitude, the three phases can be directly sampled by changing the input of P_A MUX and the reference on the comparator for Phase A is changed by REF MUX from GND to $V_{bus}/2$. ZCDx represents the zero-crossing signal of each phase. Fig. 4 illustrates this procedure:

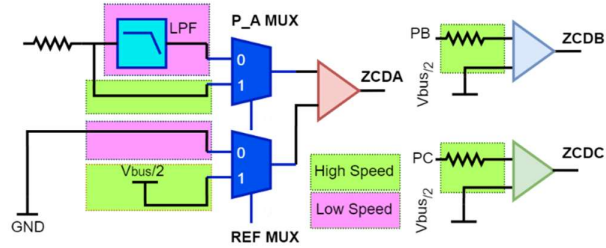

Fig. 4 BEMF sampling at low and high speeds

B. Controller Implementation

A microprocessor-based implementation of the suggested scheme is ideal for speed control. The microcontroller in the BLDC drives is responsible for receiving speed commands from the main attitude computer and transforming them into switching signals, which are passed along to the gate driving circuitry for the SSTP inverter stage of the system. The following formula can be used to determine the time between adjacent commutation periods to drive the BLDCM (using the back-EMF detection method). The frequency of computation can be found out from eq. 4 as:
$$Freq_{comp} = \left(\frac{RPM_{max}}{60}\right) * P * \phi \tag{4}$$
It should be noted that BEMF integration event happens halfway between the commutation; thus, the MCU has half the time to perform the integration operation according to eq. 5 as:
$$Freq_{intg} = 2 * Freq_{comp} \tag{5}$$
Substituting for obtained data across many robots for a maximum RPM case, the motor speed is around 5800RPM for a 2:1 force-to-weight ratio. Thus, $Freq_{comp}$ = 4060 Hz and $Freq_{intg}$ = 8129 Hz, giving us the minimum time between phase commutations and integration computation instants to be $246\mu S$ and $123\mu S$, respectively. Based on a wide choice of microcontrollers, the STM32G431 is chosen due to its high-speed clock and ADC sampling frequency, which allows a fairly wide range of PWM frequencies to be utilised with different motors since their internal electrical construction influences how fast transients can be handled.

## C. Choosing the Switching Circuitry

The Six Switch 3 -Phase Inverter (SSTP) consists of 3 half bridges and gate-driving circuitry, which transforms the low-power signals from the microcomputer to high-power switching signals necessary for driving the gate of the MOSFETs through level shifting. Fig. 5 shows a SSTP inverter:

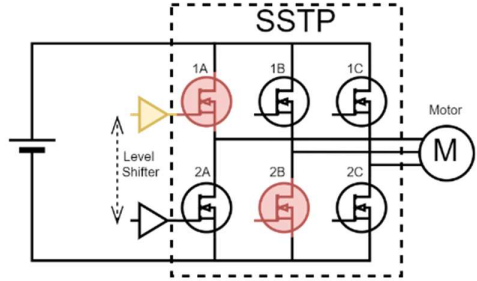

*Fig. 5 3-Phase voltage source inverter*

At an instant, as shown in Fig., when switches 1A and 2B are being conducted, the FETs act like resistive elements [7], and the RMS conduction loss is given by eq. 6:

$$P_{\text{cond}} = I_{RMS}2 \cdot R_{\text{DS(on)}} \tag{6}$$

$R_{\text{DS(on)}}$ is the on-state resistance of the FET, and $I_{I_{RMS}}$ is the RMS current being carried through the conduction channel. The formula for calculating switching power loss, which considers transistor drain-source transient switching loss and gate switching loss, is given in eq. 7:

$$P_{sw} = Psw_{rise} + Psw_{fall} + P_{gate} \tag{7}$$

Where $Psw_{rise}$ and $Psw_{fall}$ are losses from the turn-on and turn-off of the switch, proportional to the rise and fall times of the FET used and the switching frequency $F_{sw}$. $P_{gate}$ is the gate power loss, proportional with the gate charge $Q_{gate}$ and the gate driving voltage $V_{gate}$. The conduction losses are much more significant than the switching loss, and the FETS must be chosen to have the lowest possible $R_{\text{DS(on)}}$ at a compact size to ensure the solution size does not exceed space constraints. Table 1 lists some FETs with their parameters:

TABLE I

COMPARISON OF MOSFETS

| PART NO. | N CHANNEL FETS WITH $I_D$> 60A | | | |
|---|---|---|---|---|
| | $V_{DS}$, $I_D$ max. | $R_{\text{DS(on)}}$ @ $I_D$ & $V_{gate}$ | $Q_{gate}$ | Cost |
| TPN4R806PL, L1Q TOSHIBA SEMICON | 60 V, 72 A | 3.50mOhm @ 36A, 10V | 29.0 nC @ 10 V | ₹71.65000 |
| SQS160ELNW-T1_GE3 VISHAY SILICONIX | 60 V, 141 A | 4.30mOhm @ 10A, 10V | 71.0 nC @ 10 V | ₹82.48000 |
| TPH4R008QM, LQ TOSHIBA SEMICON | 80 V, 86 A | 4.00mOhm @ 43A, 10V | 57.0 nC @ 10 V | ₹107.4700 |
| PSMN1R0-30YLDX NEXPERIA | 30 V, 300 A | 2.15mOhm @ 25A, 4.5V | 57.3 nC @ 4.5 V | ₹163.2900 |
| PSMN1R2-30YLDX NEXPERIA | 30 V, 100 A | 1.60mOhm @ 25A, 4.5V | 32.0 nC @ 4.5 V | ₹129.9600 |

From the Table, it can be deduced that low $R_{\text{DS(on)}}$ losses at a target current draw are more important than the high current carrying capability of the FET, and thus, for $V_{DS} > 12V$, TPN4R806PL, L1Q is a good choice. In some cases, the battery voltage is less than 12V where logic level FETs such as the PSMN1R2-30YLDX can be used to drive the gate at 5V or lower without sacrificing a higher $R_{\text{DS(on)}}$. Switching regulators are employed to step down the battery voltage $V_{DS}$ to the gate drive voltage $V_{gate}$ to the desired level efficiently, as linear regulators would show a greater power loss in terms of heat. The gate driver's average input current is calculated and used to design the step-down circuit.

## IV. RESULTS AND ANALYSIS

The proposed control and circuitry selection method was first simulated in MATLAB/Simulink, and a custom-built electronic speed controller (ESC) was designed for hardware validation. The motor was given a step input from 400 RPM to 2000 RPM. Fig. 6 shows the behaviour of the reference speed vs. the measured speed:

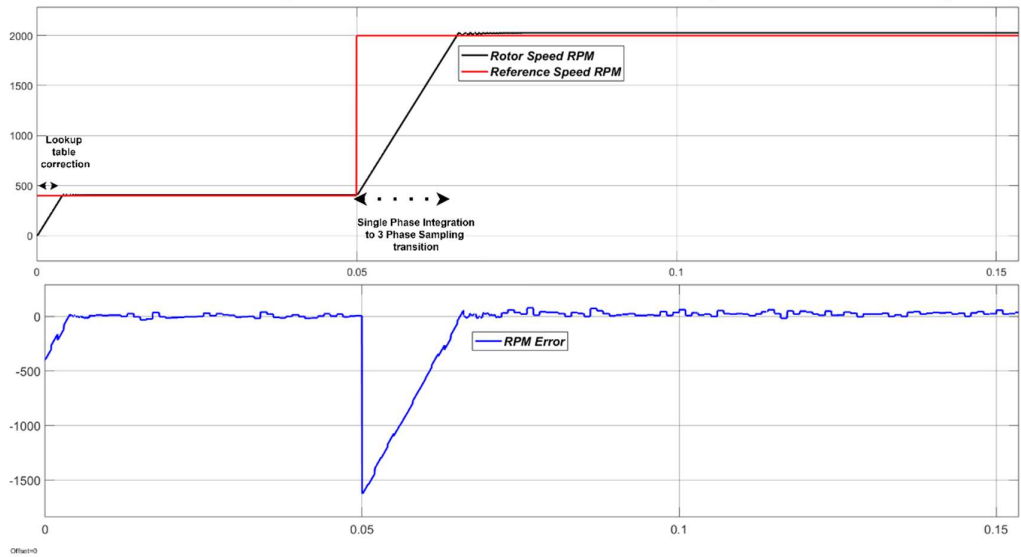

Fig. 6 Measured vs. Reference RPM error

When the controller senses back EMF, it sets the reference voltage to the circuit ground (0V) value during startup and low speeds. BEMF integration is used to enhance low-speed sensitivity and starting performance. Back EMF is identified during high-side-switch on time at higher speeds. In this region, $V_{bus}/2$ is chosen by the microcontroller as the comparator reference to detect back-EMF zero-crossing. Fig. 7 shows zero crossing detection at low speeds, and Fig. 8 shows zero crossing detection at higher speeds; it can be seen that the controller can effectively carry out phase commutation across a wide range of speed inputs.

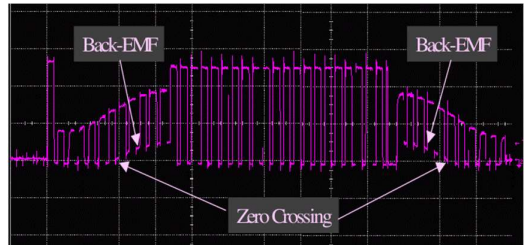

Fig. 7 Zero crossing detection at low-speeds

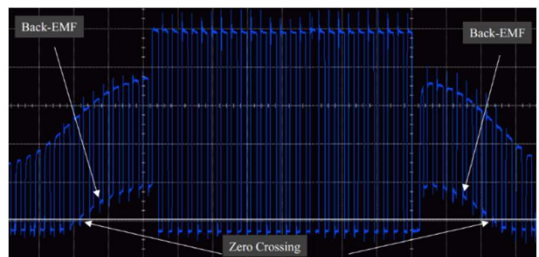

Fig. 8 Zero crossing detection at high-speeds

Three MOSFETs of similar dimensions but with different electrical parameters are shown in the table. 1 were used in building electronic speed controllers in both simulation and hardware, and their performance was validated in a drone with a Li-Po battery of nominal voltage 16V. SimScape electrical models were used in the simulation to be close to real operating characteristics. High-performance current sensors and SoC estimators were used in the hardware implementation for accurate analysis. Fig. 9 shows the plotting between simulation data and hardware testing data. The FET with lower $R_{DS\,(on)}$ gave more running time before battery depletion, which is expressed in terms of flight time (minutes).

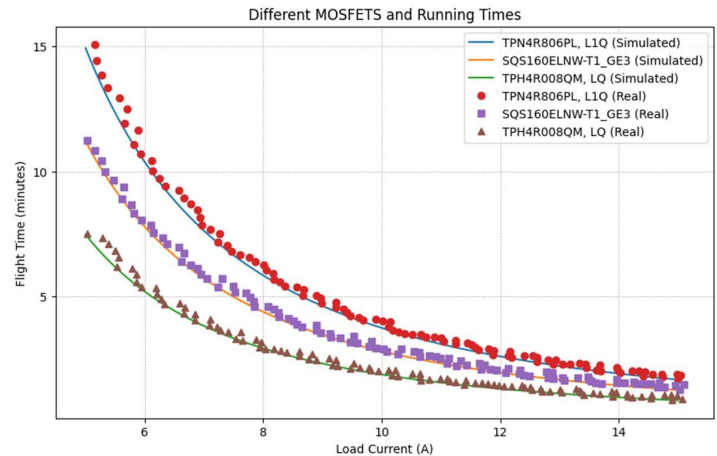

Fig 9. Effect of $R_{DS\,(on)}$ on Flight Times

## V. CONCLUSION

The proposed control scheme for brushless DC motors allows for better motor efficiency at low speeds by using the BEMF integration method. Most direct back-EMF-sensing schemes have a maximum duty-cycle limitation due to the high-side-switch minimum PWM off time at higher speeds. The improved scheme eliminates this limitation by sensing the back EMF during high-side-switch PWM on time. This combination optimises motor commutation detection over the entire speed range. The delay between switching over between the two sensing can be further improved by using higher computational power and accounting for the delay in the commutation logic. The switching elements and the gate driving circuitry are chosen to allow high efficiency, low thermal loss and longer run times.

## ACKNOWLEDGEMENT

The authors acknowledge the authors of cited papers and application notes for developing important advances in motor control strategies and express their gratitude to Texas Instruments and ST Microelectronics for hardware support in building motor drive systems.